\documentclass[letterpaper, 10pt, conference]{ieeeconf}
\IEEEoverridecommandlockouts

\usepackage{graphicx}
\graphicspath{ {figure/} }

\usepackage{cite}

\usepackage[linesnumbered,ruled,vlined]{algorithm2e}
\usepackage{bm}

\usepackage{physics}
\usepackage{amsmath}
\usepackage{mathtools}

\usepackage{amssymb} 

\usepackage{xcolor}
\usepackage{soul}
\sethlcolor{yellow}

\let\labelindent\relax
\usepackage{enumitem}
\setlist[itemize]{label=$\bullet$}

\usepackage{booktabs, tabularx, multicol, multirow}

\usepackage[noend]{algpseudocode}
\usepackage{algorithmicx}

\usepackage{color}
\usepackage{multirow}
\usepackage{booktabs}
\usepackage{url}
\usepackage{subfigure}
\usepackage{threeparttable}

\usepackage{booktabs}
\usepackage{enumerate}
\usepackage{amssymb}
\usepackage{leftidx}
\usepackage{amsfonts}
\usepackage{amsmath}
\usepackage{bm}
\usepackage{booktabs}
\usepackage{setspace}
\usepackage{makecell}
\usepackage{setspace}
\usepackage{siunitx}
\usepackage{float}

\DeclareSIUnit{\mps}{\meter\per\second}
\DeclareSIUnit{\mbps}{\mega\byte\per\second}
\DeclareSIUnit{\kbps}{\kilo\byte\per\second}
\DeclareSIUnit{\mb}{\mega\byte}
\DeclareSIUnit{\kb}{\kilo\byte}

\title{ \LARGE \bf
A Task-Agnostic, Fully-Scalable Voxel Mapping System for Large Environment
}
\title{ \LARGE \bf
A Robust, Task-Agnostic and Fully-Scalable Voxel Mapping System for Large Scale Environments
}
\author{Jinche La$^{1}$, Jun-Gill Kang$^{2}$, Dasol Lee$^{3\dagger}$%
\thanks{$^{1}$ Research Officer, {\tt\small tomcat62888@gmail.com}}%
\thanks{$^{2}$ Research Officer, {\tt\small jungillkang@gmail.com}}%
\thanks{$^{3}$ Senior Researcher, {\tt\small dasol.aero@gmail.com}}%
\thanks{All authors are with the AI Autonomy Technology Center, Agency for Defense Development(ADD), Daejeon, 34186, REPUBLIC OF KOREA.}%
\thanks{$\dagger$ Corresponding Author.}%
\thanks{* This work was supported by the Korean Government (2024).}%
}

\begin{document}
\maketitle
\thispagestyle{empty}
\pagestyle{empty}

\begin{abstract}
Perception still remains a challenging problem for autonomous navigation in unknown environment, especially for aerial vehicles. Most mapping algorithms for autonomous navigation are specifically designed for their very intended task, which hinders extended usage or cooperative task.
In this paper, we propose a voxel mapping system that can build an adaptable map for multiple tasks. The system employs hash table-based map structure and manages each voxel with spatial and temporal priorities without explicit map boundary. We also introduce an efficient map-sharing feature with minimal bandwidth to enable multi-agent applications. We tested the system in real world and simulation environment by applying it for various tasks including local mapping, global mapping, cooperative multi-agent navigation, and high-speed navigation.
Our system proved its capability to build customizable map with high resolution, wide coverage, and real-time performance regardless of sensor and environment. The system can build a full-resolution map using the map-sharing feature, with over \SI{95}{\percent} of bandwidth reduction from raw sensor data.
\end{abstract}

\section{Introduction}
\label{sec:introduction}

Drone application has significantly emerged in recent decades due to its agility, affordability, and evolution in autonomy.
One of the difficulties still remaining in drone autonomy is limited perception capability. A single drone has poor visibility in a cluttered environment due to occlusion, while it has less time to respond than a ground vehicle. Multi-agent navigation or long-range sensor may address this issue.
Recent works in multi-agent aerial navigation\cite{egoswarm}\cite{zhou2021decentralized} conducted a successful real-world experiment, but the drones still do not share perception data with other drones in the network. The potential of multi-agent system in perception task is yet to be fully utilized.
Meanwhile, most navigation and mapping algorithms still utilizes only short-range perception data despite the sensors becoming lighter and the range extended.

One of the reasons for the limited utilization of perception is data structure. Most navigation and mapping system utilizes occupancy grid map with an array-based structure.
Array-based map excels in computational efficiency, mainly due to the constant time complexity of data access. However, it lacks memory efficiency within a large environment\cite{siegwart2011introduction}. Therefore, most applications sacrifice map coverage and maintain only map data in close range. Efficiency can be reserved in this way, but application is limited to a local map of single agent. 
For application such as high-speed flight, handling long-range LiDAR data, and multi-agent task, the coverage of map should be wider while maintaining real-time performance and resolution.
Various approaches with other map have been proposed, such as Octree-based voxel map\cite{octomap} or non-voxel representations\cite{plan_voro_uuv}\cite{farplanner}. However, they often lack real-time performance or geometric details of representation.

As an attempt to address the issue of limited map utilization, we propose a voxel mapping system which can adapt to multiple types of task. It can be adopted in either local mapping, global map logging, or other applications. 
The key property of our system is robust capability in terms of maintaining all the resolution, coverage, computation, and memory efficiency at high level regardless of condition.
The proposed mapping system utilizes hash table-based data structure to maintain both appropriate resolution and coverage without sacrificing one. The system also presents a dynamic voxel management method to filter out voxels based on spatial and temporal priorities, and efficiently maintain essential voxels for its task.
Additionally, the system includes a map-sharing feature with low bandwidth of data, which can be utilized in multi-agent tasks.
Our mapping system has been tested with distinct LiDAR sensors and environments, then integrated with decentralized swarm planning algorithm EGO-SWARM\cite{egoswarm} for the demonstration of navigational applications in a simulation environment.
The contributions of this work are summarized as follows:
\begin{itemize}
\item A real-time voxel mapping system with dynamic voxel management method. It can build a highly customizable voxel map suitable for multiple types of task, with robust performance irrelevant to sensor, environment and other conditions.
\item Efficient map-sharing method for cooperative perceptional task with low data bandwidth.
\item Demonstrated the robust performance of the system and possible applications that single-purposed map cannot achieve. The demonstration was conducted using both real-world LiDAR dataset and simulation environment to show that the system can address some challenging issues in mapping and navigation task.
\end{itemize}

\section{Related Work}
\label{sec:relatedwork}

\subsection{Mapping in Autonomous Navigation}
\label{sec:rel_map}

In robotics, a role of path planning is to search for some feasible path from initial state to target state\cite{zhang2018path}.
Since the goal of this problem is to reach target state safely with efficiency, sensor data and mapping system is essential for navigation.
Grid map is a widely used map for autonomous navigation. It decomposes the space into grid and updates information of each grid, therefore robot can know information such as occupancy of arbitrary position by accessing to voxel in grid.
Some mapping system\cite{fiesta}\cite{voxblox} manipulates Euclidean Signed Distance Field (ESDF) data based on grid map, to provide useful information for obstacle avoidance.
Many planners\cite{egoplanner}\cite{raptor} save occupancy grid map in array-based data structure, mainly due to fast data access. Most other structures\cite{octomap} have worse real-time performance compared to array-based map structure. 1D array can save voxels in 3D space with proper indexing, and access to element in time complexity of $O(1)$. Also, each voxel can find its adjacent voxel by index of array without difficulty.

However, the array-based map lacks memory efficiency in a large environment, because every voxel within the designated boundary should exist in the array even if it does not have any useful information. Since the map is bounded, the whole array may need proper update when the robot moves to a new region. 
ROG-Map\cite{rogmap} utilizes circular buffer to update robocentric map without explicit destruction of voxels outside boundary, and achieved real-time performance in local mapping of high-resolution LiDAR dataset. However, it still needs separate map for additional tasks such as exploration.
Some planners use completely different representation depending on their purpose, such as Voronoi diagram\cite{plan_voro_uuv} and visibility graph\cite{farplanner}. Such representations are efficient in managing specific information, but they often lack stable performance and ability to save detailed geometric information.

Our proposed system utilizes a hash table-based data structure to achieve both high resolution and wide coverage, while an array-based map often sacrifices either resolution or map size for memory consumption. The characteristics of hash table and presented methods to improve effectiveness will be explained in following sections.

\subsection{Decentralized Multi-Agent Navigation}
\label{sec:rel_swarm}

In multi-agent navigation, each agent has to plan its trajectory in consideration of other agents. 
Many approaches such as VO(Velocity-Obstacles)-based\cite{vo}\cite{orca} and MPC(Model Predictive Control)-based\cite{dcad}\cite{dmpcp2p} have been proposed, but these approaches do not consider non-agent obstacles.
Recently, several works have proposed decentralized and asynchronous swarm planning method in dense environment, either using multi-agent reinforcement learning\cite{mapper} or trajectory optimization\cite{egoswarm}\cite{zhou2021decentralized}.
These works consider both obstacles and other agents in navigation. However, agents in these works only share trajectory data and avoid each other, which still lacks synergy of communicable multi-agent system.

Although the proposed mapping system is not a navigation system itself, it includes a map-sharing feature which can be utilized in cooperative tasks of a multi-agent group.

\subsection{Hash Table}
\label{sec:rel_hash}

A hash table stores a collection of key-value pairs with very large index set\cite{mehlhorn2008algorithms}, and searches the hash key to find its paired value. 
A hash table-based map structure has worse time complexity of $O(n)$ in worst case but better space complexity of $O(n)$ compared to array-based data structure. Therefore, it has suitable for saving data over large region.
\cite{hashreconstruction} utilizes hash table as a map container and achieved real-time performance on 3D reconstruction with $\sim$\SI{10}{\mm} resolution. Still, its focus is to accumulate sensor data and build precise reconstruction, not managing it dynamically. And the algorithm depends on GPU hardware to accelerate computation speed.
For the map application in robotics, hash table structure itself does not have any scheme to manage and remove existing voxels, unlike array-based voxel map has a spatial boundary to clear data outside of it. This can be an issue for mapping application in robotics, considering that mobile robots have relatively low computing power and memory.

To efficiently manage data and update the map in real-time, the proposed system includes dynamic voxel management method with adjustable parameters, which filters out existing voxels by spatial and temporal priorities. Also, unlike other mapping algorithms saving binary occupancy of voxel in map structure, each voxel object in the system holds information by itself to reduce search operations during update.

\section{Overview}
\label{sec:overview}

In following sections, system structure and update process are first explained in Sec. \ref{sec:mapsystem}. Then the main features of the system are summarized in Sec.\ref{sec:mapfeature}, in consideration of explanation in Sec.\ref{sec:relatedwork} and Sec. \ref{sec:mapsystem}.
Sec. \ref{sec:experiment} consists benchmark result of the mapping system using real-world LiDAR datasets. Sec. \ref{sec:application} demonstrates navigation-related applications using the proposed system.

\section{System Structure}
\label{sec:mapsystem}
\subsection{Base Structure}
\label{sec:sys_base}

\begin{figure}[b]
    \centering
    \includegraphics[trim=3cm 3.0cm 4.3cm 4.7cm, clip=true,scale=1.0 , width=\linewidth]{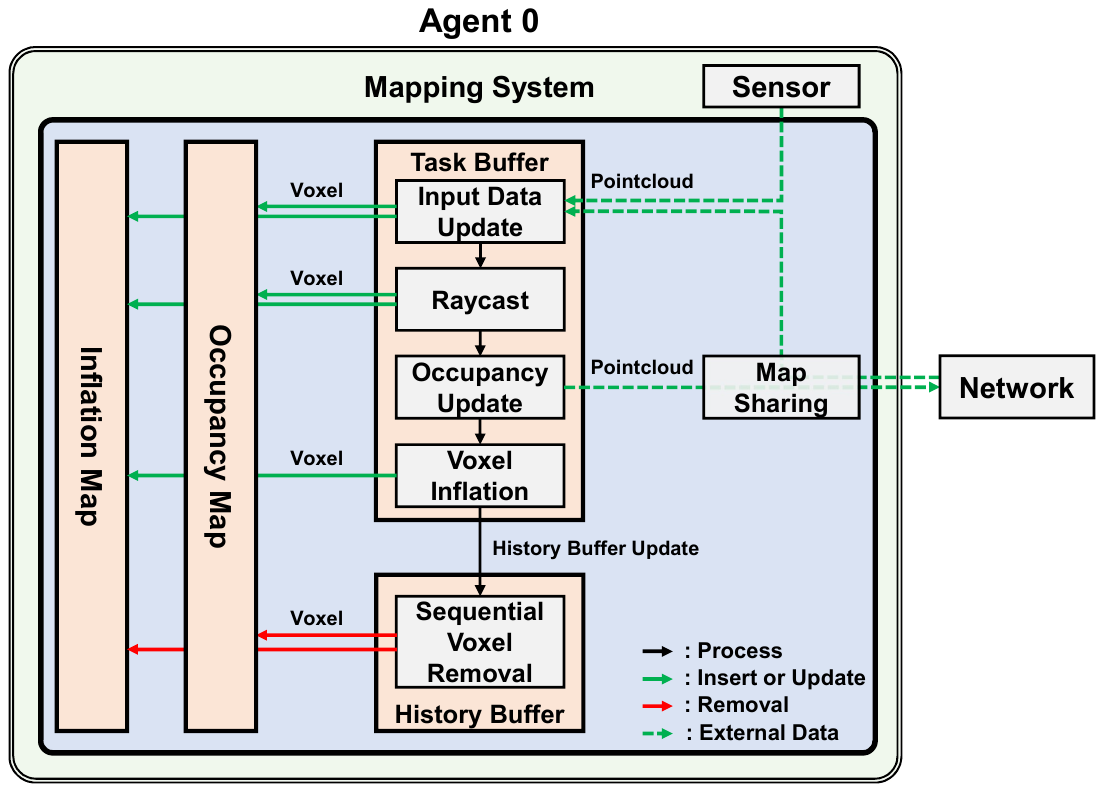}

    \caption{A block diagram of system structure.}
    \label{fig:system}
\end{figure}

Key components of system structure are map containers, task buffers, history buffer, and voxel objects. Two map containers, occupancy and inflation map, employ hash table-based data structure and represents spatial distribution of voxels. History buffer has DLL(Doubly-Linked List)-based structure and holds voxel update history sequentially. Task buffers accommodate and classify voxels during update processes. Voxel is an independent object that holds its own information, and it is mainly manipulated by pointers. 
A block diagram of system structure and update process is provided in Fig. \ref{fig:system}.

\subsection{Input Data Update Process}
\label{sec:sys_input}

\begin{algorithm}[t]
    \caption{Input Data Update}\label{algo:mapinput}
    \SetKw{KwEach}{each}
    \SetKw{KwEnv}{Notation: }
    \SetKw{KwPeriod}{every period}
    \KwEnv Input pointcloud: $\mathcal{P}$, distance between current position and $\textbf{p}_{n}$: $d_n$. \\
    
    \For{$ \textbf{p}_{n} \in \mathcal{P} $}{
        \If{$d_n > d_{in}$}{$return$\;}
        $ \textbf{i}_n \gets \textbf{posToKey}(\textbf{p}_n) $\;
        \If{$ \textbf{i}_n \notin \mathcal{B}_{new}$ \textbf{and} $\textbf{p}_n \notin \mathcal{B}_{keep} $}{
            \uIf{$ \textbf{i}_{n} \notin \mathcal{M}_{occ} $}{
                \uIf{$ \textbf{i}_{n} \notin \mathcal{M}_{inf} $}{
                    $ vox_{n,p} \gets \textbf{makeVoxel}(\textbf{i}_{n}) $\;
                    $ vox_{n,p}.\textbf{insertTo}(\mathcal{B}_{new}, \mathcal{M}_{occ}, \mathcal{M}_{inf}) $\;
                    $vox_{n}.l_{occ} += \textbf{logit}(p_{init})$\;
                }
                \Else{
                    $ vox_{n,p} \gets \mathcal{M}_{inf}.\textbf{findVox}(\textbf{i}_{n}) $\;
                    $ vox_{n,p}.\textbf{insertTo}(\mathcal{B}_{new}, \mathcal{M}_{occ}) $\;
                }
            }
            \Else{
                $ vox_{n,p} \gets \mathcal{M}_{occ}.\textbf{findVox}(\textbf{i}_{n}) $\;
                $ vox_{n,p}.\textbf{insertTo}(\mathcal{B}_{keep}) $\;
                \uIf{$ \textbf{i}_n \notin \mathcal{M}_{inf} $}{
                    $ vox_{n,p}.\textbf{insertTo}(\mathcal{M}_{inf}) $\;
                }
            }
        }
        $ \textbf{raycastProcess}(vox_{n,p}, pos_{self}) $\;
    }
\end{algorithm}

\begin{figure}[t]
  \centering
\includegraphics[trim=8cm 2.8cm 1cm 2.6cm, scale=1.0, clip=true, width=\linewidth ]{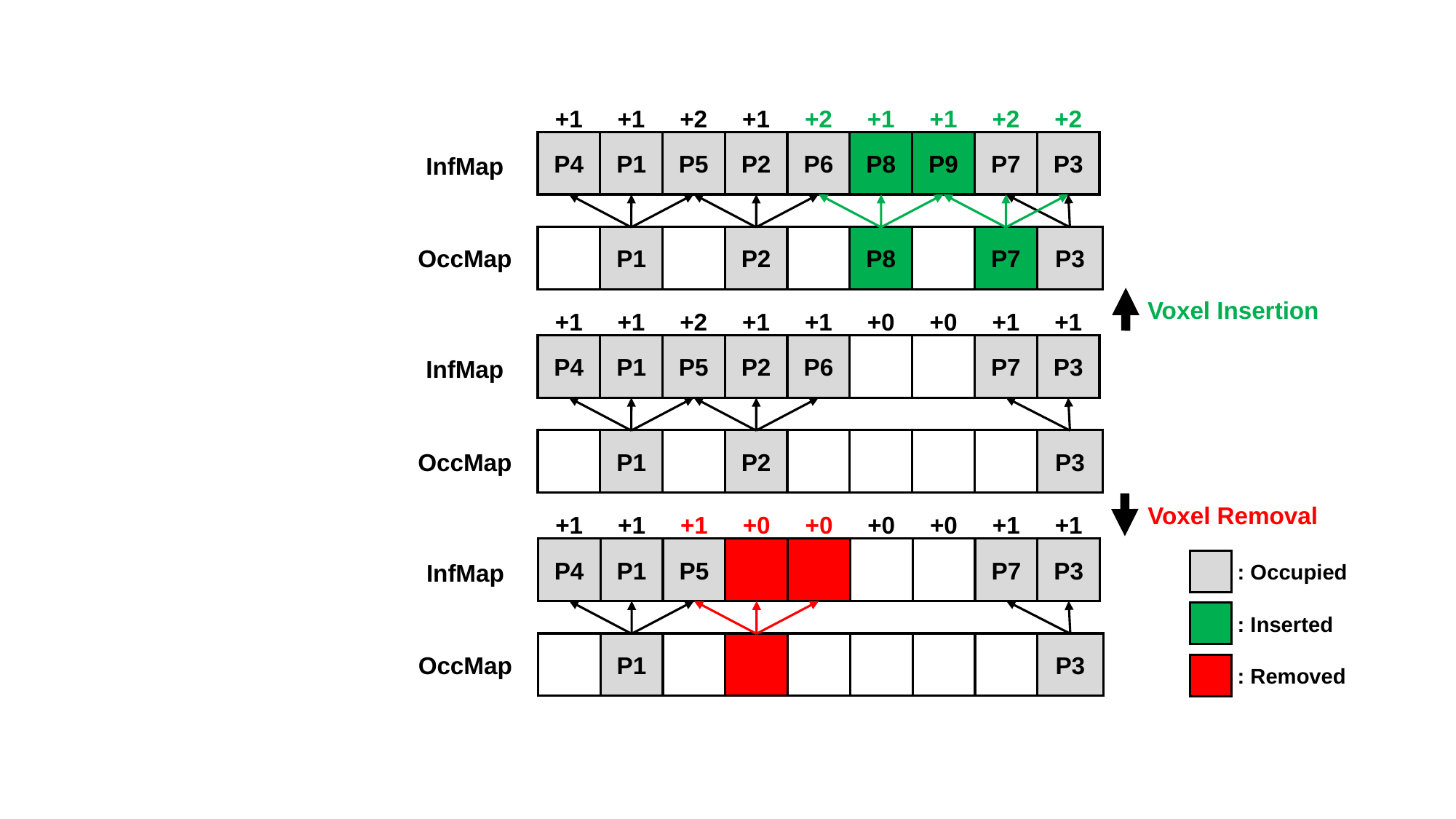}
  \caption{Example of pointer-based voxel insertion and removal in 1D map structure. A pointer with the same number refers to the same voxel object.}
  \label{fig:voxelpointer}
\end{figure}

Process of input data update is presented in \textbf{Algorithm \ref{algo:mapinput}}. The process utilizes two map containers $\mathcal{M}_{occ}$, $\mathcal{M}_{inf}$ and two task buffers $\mathcal{B}_{new}$, $\mathcal{B}_{keep}$. For a point $\textbf{p}_n = (p_x, p_y, p_z)$, the corresponding key $\textbf{i}_n = (i_x, i_y, i_z)$ is defined as

\begin{equation} \label{eq:key_index}
    \textbf{i}_n = floor((\textbf{p}_n-\textbf{o})/res) 
\end{equation}

where $res$ is resolution and map origin is $\textbf{o} = (o_x, o_y, o_z)$. Instead of a structural boundary, the process can choose to discard a point outside the input distance threshold $d_{in}$ (Line 3-4). 
For a point $\textbf{p}_n$, a voxel pointer $vox_{n,p}$ makes a pair of hash key $\textbf{i}_n$ and self as a hash value. For each point in input pointcloud, existence of the hash key-value pair is searched within $\mathcal{B}_{new}$ and $\mathcal{B}_{keep}$ (Line 6). If the pair is not in both buffers, a voxel pointer $vox_{n,p}$ is made or copied to map container and inserted to either task buffer (Line 7-22). Insertion or removal of voxel pointer in map is shown in Fig. \ref{fig:voxelpointer} with 1D example. The system shares voxel object between two maps by copying pointer for synchronization. 

\subsection{Occupancy Check Process}
\label{sec:sys_occ}

Occupancy check process is based on the probabilistic approach presented in \cite{moravec1985high}, and checks voxels in task buffers. Logit probability of occupancy $\textbf{L}(n\mid s_{1:t})$ of voxel $vox_n$ can be estimated by sensor measurement history $s_{1:t}$ and logit probability of current step $L(n \mid s_{t})$.

\begin{equation} \label{eq:logprob}
L(n \mid s_{1:t}) = L(n \mid s_{1:t-1}) + L(n \mid s_{t})
\end{equation}

Logit probability of occupancy in the current step $L(n \mid s_{t})$ is calculated from hit count and miss count. 

\begin{equation} \label{eq:logprob2}
L(n \mid s_{t}) = n_{hit}l_{hit} + n_{miss}l_{miss}
\end{equation}

Hit count increases if the input point $\textbf{p}_n$ matches the voxel, and miss count increases if the ray passes the voxel during raycast process. Logit probability multipliers for hit and miss are $l_{hit}$ and $l_{miss}$.
Occupancy probability is clamped between $l_{min}$ and $l_{max}$ and occupancy state is either \texttt{Occ}, \texttt{Unknown}, or \texttt{Free} based on threshold $l_{occ}$ and $l_{free}$. After occupancy check, the process flags voxels that occupancy state has been changed to or from \texttt{Occ} in current step as \texttt{Occ-changed}. The system only checks the voxels with \texttt{Occ-changed} flag in occupancy inflation process to reduce computation. For memory efficiency, known-free voxels are inserted to $\mathcal{B}_{del}$ and removed from the map containers in the sequential voxel removal process.

\subsection{Occupancy Inflation Process}
\label{sec:sys_inf}

\begin{algorithm}[t]
    \caption{Occupancy Inflation}\label{algo:inflation}
    \SetKw{KwEach}{each}
    \SetKw{KwEnv}{Notation: }
    \SetKw{KwPeriod}{every period}
    \KwEnv inflating voxel pointer: $vox_{check,p}$, inflated voxel pointer: $vox_{inf,p}$, inflation look-up table: $\mathcal{T}_{inf}$, inflation count: $n_{i}$, key list to inflate: $\mathcal{G}_{inf}$.\\
    \If{$\textbf{getDist}(vox_{check,p}, pos_{self}) < d_{inf} $ and $ vox_{check,p}.\textbf{isOccChanged}()$}{
        \For{$vox_{inf,p} \in vox_{check,p}.\mathcal{T}_{inf}$}{
            \If{ not $vox_{check,p}.\textbf{isOcc}()$}{
                $ vox_{inf,p}.\textbf{removeFrom}(vox_{check,p}.\mathcal{T}_{inf}) $\;
                
                $ vox_{inf,p}.n_{i} \gets vox_{inf,p}.n_{i} - 1 $\;
            }
        }
        \For{$ vox_{inf,p} \notin vox_{check,p}.\mathcal{T}_{inf} $ and $ vox_{inf,p} \in \mathcal{G}_{inf} $}{
            \uIf{ not $vox_{check,p}.\textbf{isOcc}()$}{
                $ vox_{inf,p}.\textbf{removeFrom}(vox_{check,p}.\mathcal{T}_{inf}) $\;
                $ vox_{inf,p}.n_{i} \gets vox_{inf,p}.n_{i} - 1 $\;
            }
            \ElseIf{ $vox_{check,p}.\textbf{isOcc}()$}{
                $ vox_{inf,p}.\textbf{insertTo}(vox_{check,p}.\mathcal{T}_{inf}) $\;
                $ vox_{inf,p}.n_{i} \gets vox_{inf,p}.n_{i} + 1 $\;
            }
        }
    }
\end{algorithm}

Our mapping system presents an inflation look-up table for each voxel object to search other voxels efficiently. 
Inflation look-up table of $vox_{check}$ contains voxel pointers that are inflated by $vox_{check}$. The inflation process searches target voxel pointer $vox_{inf,p}$ (Line 3,7) and updates look-up table of $vox_{check}$ and state of $vox_{inf}$ based on occupancy (Line 4-6, 8-13). After initial inflation, each voxel can simply access to inflated voxels through the look-up table (Line 3).
Additionally, the inflation process only updates the voxels with \texttt{Occ-changed} flag and distance threshold $d_{inf}$(Line 2) to block unnecessary computation on stable or far-away voxels.

\subsection{Sequential Voxel Removal Process}
\label{sec:sys_rem}

\begin{algorithm}[t]
    \caption{Sequential Voxel Removal}\label{algo:memory}
    \SetKw{KwEach}{each}
    \SetKw{KwEnv}{Notation: }
    \SetKw{KwPeriod}{every period}
    \KwEnv history buffer: $\mathcal{B}_{his}$, voxel limit size: $n_{lim}$, iterator of history buffer: $itr$.\\
    \For{$ vox_{n,p} \in \mathcal{B}_{new} $}{
        \uIf{$ vox_{n,p}.\textbf{isOcc}() $}{
            $ vox_{n,p}.\textbf{pushBackTo}(\mathcal{B}_{his}) $\;
            $ vox_{n,p}.itr \gets \textbf{getItr}(\mathcal{B}_{his}.\textbf{back}()) $\;
        }
        \ElseIf{$ vox_{n,p}.\textbf{isToRemove}()$}{
            $ vox_{n,p}.\textbf{insertTo}(\mathcal{B}_{del}) $\;
        }
    }
    \For{$ vox_{n,p} \in \mathcal{B}_{keep} $}{
        $ \mathcal{B}_{his}.\textbf{erase}(vox_{n,p}.itr) $\;
        \uIf{$ vox_{n,p}.\textbf{isOcc}() $}{
            $ vox_{n,p}.\textbf{pushBackTo}(\mathcal{B}_{his}) $\;
            $ vox_{n,p}.itr \gets \textbf{getItr}(\mathcal{B}_{his}.\textbf{back}()) $\;
        }
        \ElseIf{$ vox_{n,p}.\textbf{isToRemove}()$}{
            $ vox_{n,p}.\textbf{insertTo}(\mathcal{B}_{del}) $\;
        }
    }
    $ \textbf{removeElemFromMap}(\mathcal{B}_{del}) $\;
    \While{$ \textbf{sizeOf}(\mathcal{B}_{his}) > n_{lim} $}{
        $ vox_{n,p} \gets \mathcal{B}_{his}.\textbf{front}() $\;
        $ \mathcal{B}_{his}.\textbf{pop\_front}() $\;
        $ \textbf{removeFromMap}(vox_{n,p}) $\;
    }
\end{algorithm}

As a part of dynamic voxel management, the system introduces sequential voxel removal process. The process automatically removes old voxels to limit the number of voxels, as shown in \textbf{Algorithm \ref{algo:memory}}. For a new voxel pointer in $\mathcal{B}_{new}$ (Line 2), the pointer is also inserted in history buffer (Line 4) and the corresponding iterator is saved inside voxel object (Line 5). For a voxel pointer in $\mathcal{B}_{keep}$, the process removes the pointer from history buffer by iterator (Line 9) and inserts again if the voxel is occupied (Line 10-12). Therefore, history buffer stores recently inserted or updated voxels in its rear part, and the old voxels will gradually move towards the front of the history buffer. After deleting known-free voxels in $\mathcal{B}_{del}$ (Line 15), the process removes voxel in front of history buffer until map size reaches below its limit (Line 16-19). Time complexity of history buffer-related process is $O(1)$ due to the characteristic of DLL.

\section{System Feature}
\label{sec:mapfeature}

This section provides summary and supplementary explanation of main features introduced in the mapping system.

\subsection{Pointer-Based Voxel Manipulation}
\label{sec:feature_pointer}

Array-based occupancy grid map often maintains individual arrays for each property, such as occupancy and raycast counter. 
However, maintaining multiple hash table would result in worse performance due to its time complexity. Therefore, our system organizes each voxel as an individual object containing information such as hash key, inflation look-up table, occupancy probability, etc. The system builds only a single voxel object per hash key and distributes it using multiple pointers for efficiency. 
In this way, any update in a voxel can be immediately disseminated to map containers and buffers. Also, each voxel can update inflation state using pointers stored in the inflation look-up table.

\subsection{Dynamic Voxel Management}
\label{sec:feature_filter}

\begin{figure}[b]
  \centering
\includegraphics[trim=0cm 6.5cm 0cm 14.2cm, scale=1.0, clip=true, width=\linewidth ]{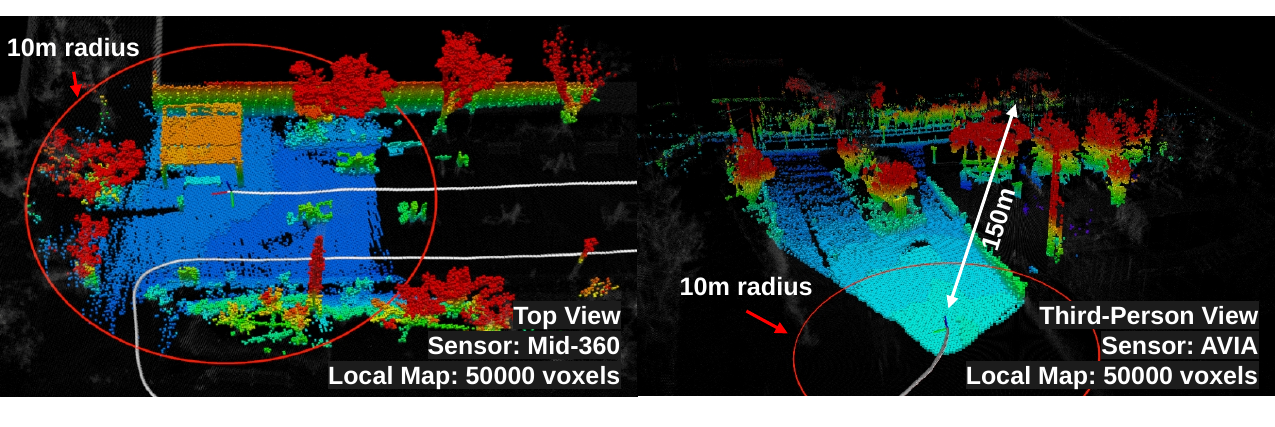}
  \caption{Left: Local map using Livox Mid-360 LiDAR. Right: Local map using Livox Avia LiDAR with same parameters. Colored region represent local map. The maps are vastly different, because the map is automatically adjusted according to sensor data.}
  \label{fig:localmap}
\end{figure}

Our system presented the voxel management method with spatial and temporal priorities because hash table-based data structure does not have any inherent criteria to filter out unnecessary voxels.
The method can choose to limit the range of input and inflation for spatial control and filter out the old voxels for temporal control, as seen in previous section.
Consequently, the management method enables active control of memory consumption and map data. The system can build \textbf{non-robocentric local map} using temporal control, as shown in Fig. \ref{fig:localmap}. It can automatically maintain awareness where it is needed, based on incoming sensor data and parameters.

\subsection{Map-Sharing Feature}
\label{sec:feature_share}
\begin{figure}[t]
  \centering
  \includegraphics[trim=1cm 2.9cm 1cm 6.5cm, scale=1.0, clip=true, width=\linewidth ]{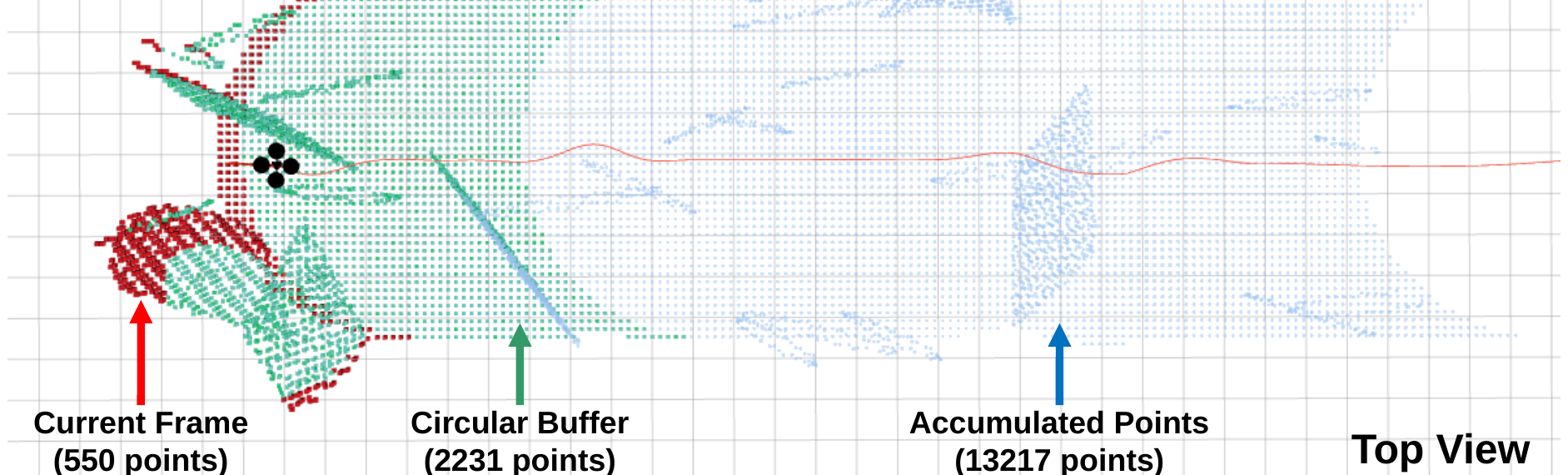}
  \caption{Comparison between broadcast of single frame, data within circular buffer, and whole accumulated data.}
  \label{fig:circularbuffer}
\end{figure}

Our system is suitable for utilizing multiple sensor data. However, sharing sensor data or whole map involves excessive computation and bandwidth consumption. Instead, a lightweight map-sharing feature is presented, which saves and exports newly occupied voxels.
Communication status should be considered to further optimize the export data. Discarding past frames during connection loss and broadcasting only the latest frame would make a shared map fragmented. In contrast, accumulating every frame during connection loss is not efficient due to inconsistent size of payload.
To compromise between two and transmit the data efficiently, the map-sharing feature utilizes circular buffer to save recent frames. It can automatically maintain its size and continuity of data by implicitly overwriting old data with a recent one. Comparison of these approaches is visualized in Fig. \ref{fig:circularbuffer}. 
Received shared data goes through the same process as onboard sensor data except for raycast and map export process. 

\subsection{Parameter adjustment}
\label{sec:feature_param}

The user can dynamically adjust key parameters without invalidating saved information. 
Parameters such as initial occupancy probability $p_{init}$, voxel limit $n_{lim}$, input range $d_{in}$, and inflation range $d_{inf}$ can be modified during operation, and will affect after current cycle of update. 
Several applications are possible with combination of parameters. For example, 1) local mapping: limits input range, inflation range, and voxel limit based on sensor and speed. 2) autonomous mapping: increase input range and limit inflation range for autonomous navigation and global mapping. 3) passive perception: with or without sensor, increase input and inflation range, receive map data from nearby robots to enhance awareness.

\section{Benchmark}
\label{sec:experiment}

\subsection{System Performance}
\label{sec:exp_map}

\begin{table*}
\begin{threeparttable}[t]
\scriptsize
\centering
\caption{Benchmark Comparison}
\label{tab:benchmark}
\begin{tabularx}{\textwidth}{@{}l *{13}{c} @{}} 
\toprule

& \multicolumn{6}{c}{Parameter} & \multicolumn{7}{c}{Result} \\ \cmidrule(r){1-6} \cmidrule(r){7-13} 
Setup&Sensor& $f_\mathrm{upd}$ (Hz)& $d_\mathrm{in}$ (m) & $d_\mathrm{inf} (m)$  & $n_\mathrm{lim}$ & $n_\mathrm{occ}$ & $n_\mathrm{inf}$ & $t_\mathrm{tot}$ (ms) & $t_\mathrm{occ}$ (ms) & $t_\mathrm{inf}$ (ms) & $t_\mathrm{m}$(ms) & $m_{max}$ (MB)\\ 
\midrule

\textbf{Park\tnote{1}-a}& self &  10 & Inf & 8 & 50000 & 16837 & 114970.585 & 9.046 & 0.464 & 4.599 & 2.556 & 166\\
\textbf{Park-b}& self &  10 & Inf & 8 & 250000 & 119847 & 495029.903 & 8.897 & 0.545 & 4.646 & 1.915 & 445\\
\textbf{Park-c}& self &  10 & Inf & 8 & 850854\tnote{4} & 494166 & 3576456.505 & 8.671 & 0.307 & 6.581 & 0.000 & 1220\\
\textbf{Park-d}& relayed\tnote{5} &  10 & Inf & Inf & 50000 & 50000 & 986485.866 & 18.691 & 0.056 & 13.527 & 4.613 & 351\\
\textbf{Park-e}& self &  10 & 10 & 8 & 50000 & 29196 & 226604.243 & 8.859 & 0.205 & 5.280 & 2.485 & 185\\

\midrule

\textbf{Forest\tnote{2}-a}& self &  10 & Inf & 8 & 50000 & 8446 & 95507.048 & 9.002 & 0.435 & 4.646 & 2.594 & 148\\
\textbf{Forest-b}& self &  10 & Inf & 8 & 250000 & 68449 & 345307.508 & 9.170 & 0.561 & 4.598 & 2.255 & 327\\
\textbf{Forest-c}& self &  10 & Inf & 8 & 2116571\tnote{4} & 528941 & 5394112.776 & 9.710 & 0.488 & 7.009 & 0.000 & 1311 \\
\textbf{Forest-d}& relayed\tnote{5} &  10 & Inf & Inf & 50000 & 50000 & 1554734.021 & 18.242 & 0.052 & 14.196 & 3.578 & 422  \\
\textbf{Forest-e}& self &  10 & 10 & 8 & 50000 & 16544 & 228970.333 & 9.064 & 0.224 & 5.704 & 2.361 & 179\\

\midrule

\textbf{Urban\tnote{3}-a}& self &  40 & Inf & 8 & 50000 & 48174 & 5417.848 & 1.499 & 0.279 & 0.209 & 0.324 & 67 \\
\textbf{Urban-b}& self &  40 & Inf & 8 & 250000 & 249112 & 25175.199 & 1.799 & 0.349 & 0.231 & 0.278 & 170 \\
\textbf{Urban-c}& self &  40 & Inf & 8 & 2205290\tnote{4} & 2204109 & 159948.172 & 1.457 & 0.319 & 0.112 & 0.000 & 741 \\
\textbf{Urban-d}& relayed\tnote{5} & 10 & Inf & Inf & 50000 & 50000 & 32951.897 & 18.346 & 0.151 & 13.733 & 3.193 & 410 \\
\textbf{Urban-e}& self &  40 & Inf & 400 & 50000 & 49578 & 455590.646 & 9.076 & 0.076 & 7.297 & 1.702 & 510  \\

\bottomrule 
\end{tabularx}
    \begin{tablenotes}
    \item [1] {Livox Mid-360 360{\textdegree} LiDAR. Average input points per frame: 4027, Sensor Rate: \SI{10}{\hertz}, Scene size: \SI{100}{\meter}$\times$\SI{63}{\meter}$\times$\SI{12}{\meter}, Travel distance: \SI{280.26}{\meter}.}
    \item [2] {Ouster OS0-32 360{\textdegree} LiDAR. Average input points per frame: 3067. Sensor Rate \SI{10}{\hertz}, Scene size: \SI{200}{\meter}$\times$\SI{85}{\meter}$\times$\SI{16}{\meter}, Travel distance: \SI{214.25}{\meter}.} 
    \item [3] {Livox Avia solid-state LiDAR. Average input points per frame: 3340. Sensor Rate: \SI{40}{\hertz}, Scene size: \SI{450}{\meter}$\times$\SI{375}{\meter}$\times$\SI{160}{\meter}, Travel distance: \SI{332.45}{\meter}.}
    \item [4] {Experiment without voxel limit. $n_\mathrm{lim}$ indicates total voxels stored in the occupancy map for this case.} 
    \item [5] {Received data from another system using map-sharing feature instead of sensor data.} 
    \end{tablenotes}
\end{threeparttable}%
\end{table*}

\begin{figure*}[p]
    \centering
    \includegraphics[trim=0cm 13.0cm 0cm 0cm, scale=1.0, clip=true, width=1.0\textwidth ]{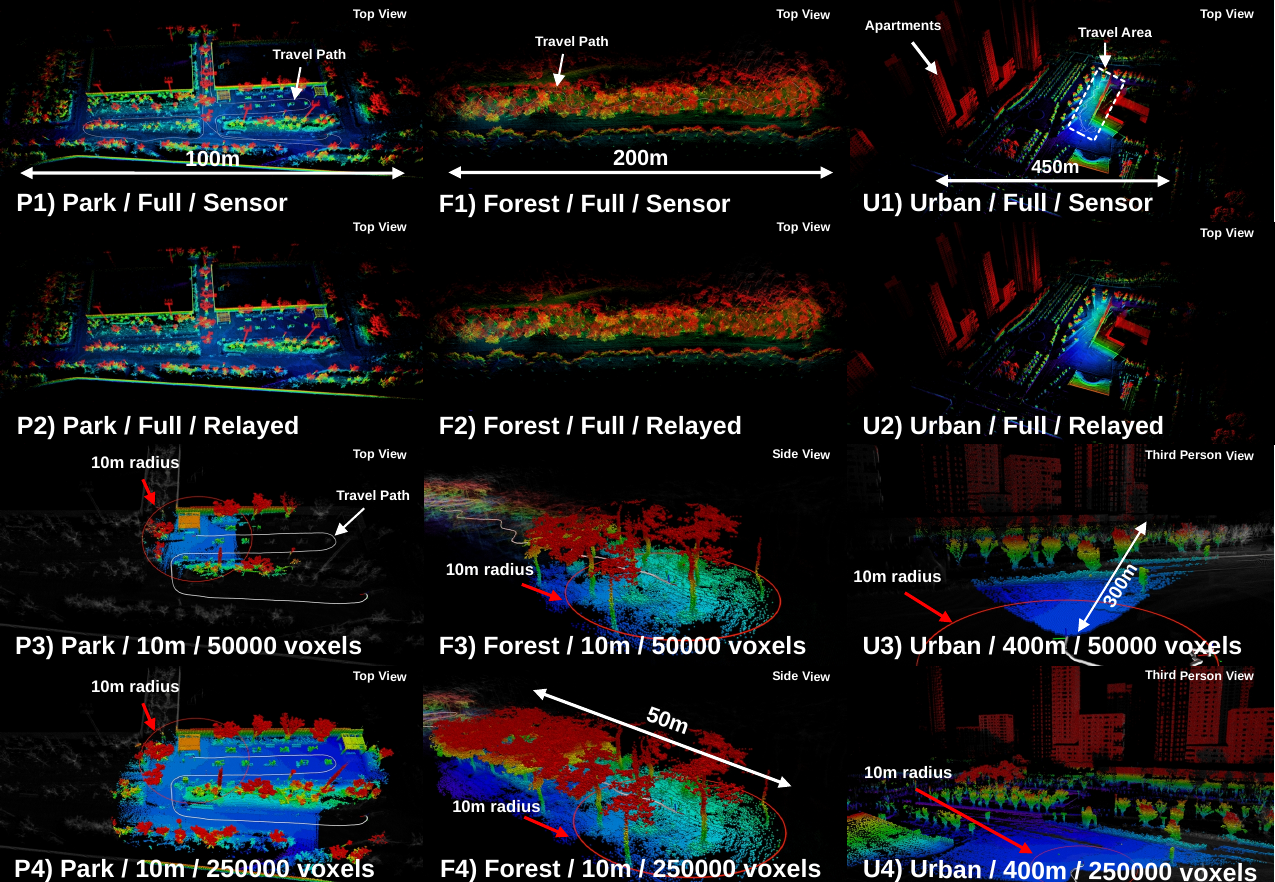}
    \caption{Visualized result of mapping experiment from distinct LiDAR datasets. Red curves indicate traveled path. Distance written in tag indicates input range, and pts indicates voxel limit.}
    \label{fig:maptest}
    \vspace{\floatsep} 
    
    \centering
    \includegraphics[trim=0.0cm 0.0cm 0.0cm 16.5cm, scale=1.0, clip=true, width=\textwidth ]{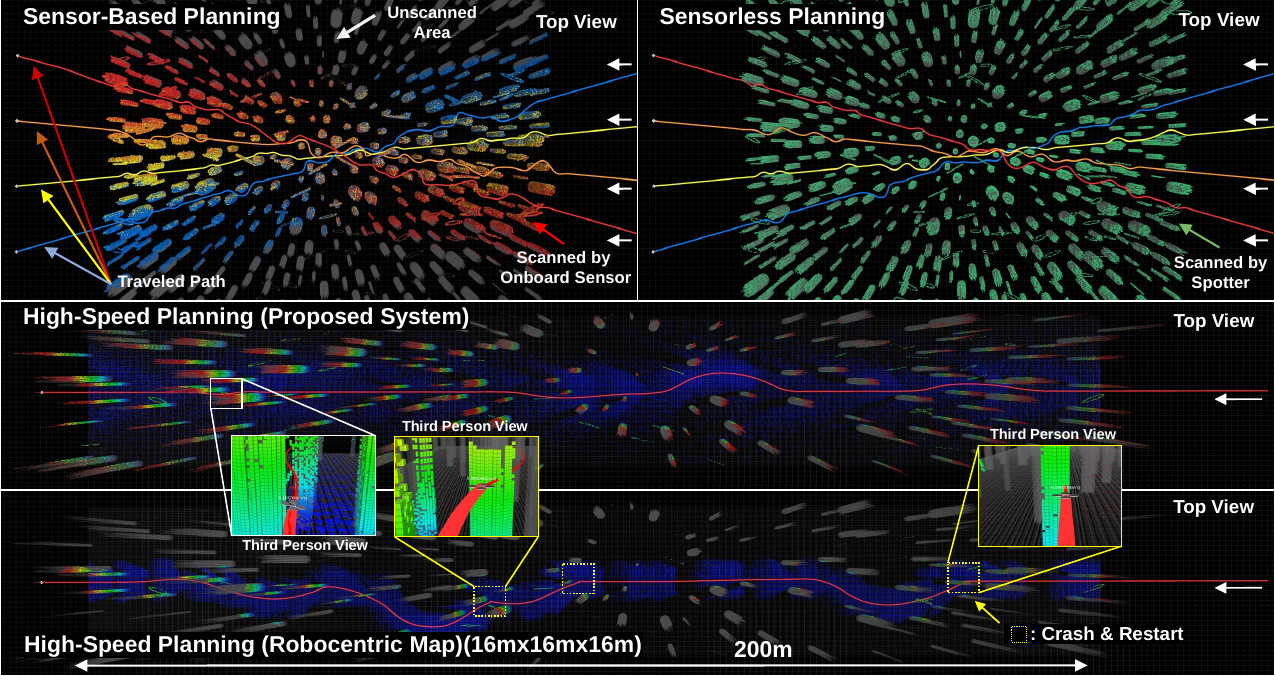}
    \caption{Up: Comparison of navigation scenarios. Upper Left: Result of onboard sensor-based navigation. Each colored region indicates drone's sensor data. Upper Right: Result of sensorless navigation. Green color indicates shared map data from $spotter$ drone.
    Down: Comparison of high-speed navigation result in top view. Robocentric map failed to provide enough time margin for planner to react in higher speed. }
    \label{fig:application}
\end{figure*}

In this section, we tested the characteristics of our system on three datasets: \textbf{Park}, \textbf{Forest}, and \textbf{Urban}. \textbf{Park} is a small park environment with moderately dense structure such as walls, streetlights, and bushes, and we scanned the environment using Livox Mid-360 LiDAR. 
\textbf{Forest} is a sparsely occupied outdoor environment under the canopy of pine trees, and we scanned the environment using Ouster OS0-32 LiDAR. 
\textbf{Urban} is a city park with large buildings are at the background. We used solid-state LiDAR sensor Livox Avia with maximum detection range of \SI{450}{\meter} to scan the scene. 
Detailed information about the datasets including scene size are written int the Table \ref{tab:benchmark}
. For every experiment, obstacle inflation distance is \SI{0.2}{\meter}.
Mapping resolution is \SI{0.1}{\meter} for Park and Forest. For Urban, mapping resolution is $0.2(m)$ and $p_{init}$ is set higher to quickly capture long-range points. All dataset utilized LiDAR inertial odometry algorithm FAST-LIO\cite{fastlio}.
The computing unit for benchmark is a laptop with Ryzen 5900HS mobile CPU, which has approximately \SI{75}{\percent} performance of Intel i5-1340P in NUC 13 Pro.

Table \ref{tab:benchmark} presents the benchmark result of experiments. 
We compare the results: number of occupied voxels $n_\mathrm{occ}$, average number of inflated voxels $n_\mathrm{inf}$, total computation time $t_\mathrm{tot}$, inflation processing time $t_\mathrm{inf}$, voxel management processing time $t_\mathrm{m}$, and maximum memory consumption $m_\mathrm{max}$.
In overall, the proposed mapping system maintained stable real-time performance across various setups. Every processing time of sensor data took less than \SI{10}{\ms}, including 40Hz data of \textbf{Urban} dataset.
In terms of efficiency, the proposed system used \num{8.509E5} voxels to build a full-scene map in Park dataset, while an array-based map would require \num{7.560E7} voxels with the same resolution. The comparison for Forest dataset results in \num{2.117E+06} and \num{2.720E+08} voxels each. The system used \num{2.205E+06} voxels for complete mapping of Urban environment while the array-based map requires \num{3.375E+09} voxels, which is more than 1000 times of the proposed mapping system. 

Visualized map data are shown in Fig. \ref{fig:maptest}. 
Images with 'Full' tag visualizes global map stored within the system (Setup c, d). Comparison between the maps using sensor data (Image P1,F1,U1) and the maps using map-sharing feature (Image P2, F2, U2) shows that our system can construct precise map by utilizing map-sharing feature. The 'Relayed' maps maintained \SI{99.5}{\percent} voxels compared to original 'Sensor' maps in average, with same resolution.
Other images show that the proposed system can automatically adjust its local map coverage, based on sensor configuration and parameters. For example, Two local maps (Image U3, U4) in \textbf{Urban} dataset form circular sector-shape with several hundred meters size. Because the sensor has limited FOV(Field-Of-View), the local map does focus on covering the frontal area where sensor data are concentrated, results in non-robocentric map. Local map is relatively robocentric With 360{\textdegree} LiDAR sensor (Image P3, P4, F3, F4), and it can preserve old records for a longer period depending on $n_{lim}$(Image P4, F4).

The system has proved that it can maintain high resolution, wide coverage, and real-time performance regardless of the condition.
The potential issue in the result is that obstacle inflation consumes most of the processing time and memory. Although total process cycles in real-time, inflation process still takes quite a portion of time despite the additional methods presented by the system to improve searching efficiency of voxels. This time-consuming process also affects map-sharing, because receiving shared map data with unlimited inflation range causes full inflation once the data is received.

\subsection{Effect of Map-Sharing}
\label{sec:exp_mapsharing}

Presented map-sharing feature in mapping system can relay the map data with massively reduced bandwidth but without downsampling or resolution loss.
The average bandwidth of original sensor data is \SI{1.86}{\mbps}, \SI{1.91}{\mbps}, and \SI{7.42}{\mbps} for Park, Forest, and Urban dataset. Bandwidth of each shared data is \SI{57.0}{\kbps}, \SI{60.1}{\kbps}, and \SI{131.5}{\kbps}, which is \SI{97.3}{\percent} reduction in average size from sensor data.

\subsection{Effect Of Voxel Pointers}
\label{sec:exp_pointer}

To check the effect of pointer management, a separate test was conducted using Park dataset. When \num{100000} voxel pointers were stored within occupancy map, memory consumption was \SI{75.4}{\mb}. When inflation was applied only in position of each voxel, memory consumption was changed to \SI{77.8}{\mb}. The small difference indicates that maintaining multiple pointers in maps and buffers for a single voxel does not make significant overhead. 

\section{Applications}
\label{sec:application}

We conducted several demonstrations on simulation environment to show that our mapping system is applicable for complicated navigational tasks.

\subsection{Planner Integration and Simulation}
\label{sec:app_planner}
We integrated our system with multi-agent planner \textbf{EGO-Swarm} for demonstration. The simulation environment and dynamics are based on its simulator code. We converted the base simulator to ROS 2 environment and added some functionalities including simulated LiDAR sensor and communication.

\subsection{Sensorless Navigation}
\label{sec:app_sensorless}

In this scenario, a group of drones performs autonomous navigation in a dense simulation environment without onboard sensor. 
Two $spotter$ drones stay in higher altitude to broadcast its map data to other $traveler$ drones. The sensor configuration of $spotter$ drones is \SI{30}{\meter} range, \SI{100}{\degree} HFOV(Horizontal Field-Of-View), \SI{60}{\degree} VFOV(Vertical Field-Of-View), \SI{5}{\hertz} frequency and \SI{80}{\degree} pitched sensor pose towards the surface. 
The sensor configuration of onboard sensor-based navigation scenario for comparison is \SI{10}{\meter} range, \SI{360}{\degree} HFOV, \SI{90}{\degree} VFOV and \SI{10}{\hertz} frequency. The velocity limit of the planner is \SI{3}{\mps} and planning horizon is \SI{9.5}{\meter}. The test involved 10 iterations. 

The result in Fig. \ref{fig:application} (upper part) and Table \ref{tab:sensorless} shows the comparison of \textbf{1) onboard sensor-based navigation} and \textbf{2) sensorless navigation based on map-sharing feature}. 
The paths in Fig. \ref{fig:application} show that navigation using shared map data can evade obstacles as well as onboard sensor-based navigation can. 
Table \ref{tab:sensorless} shows that sensorless navigation achieved better navigation result without collision, because $spotter$ drone can scan obstacles that cannot be seen from $traveler$ drones. 
Bandwidth of map-sharing in sensorless scenario was \SI{105.2}{\kbps} combined, and \SI{3.58}{\mbps} combined for sensor data in sensor-based scenario.
The demonstration shows that the vehicles with different hardware can cooperate in navigation with each specialized task. It can also work in different combinations, or hybrid of map-sharing and onboard sensor. Our system can easily perform multi-agent task without issue in utilizing remote map data.

\begin{table}[b]
	\centering
	\caption{ Result Comparison of Sensorless Navigation Scenario}  
        \begin{tabularx}{\linewidth}{{c|cccc}}
		\toprule
		Setup & $d_{\rm{fly}}$(m) &  $t_{\rm{fly}}$(s) & $t_{\rm{cal}}$(ms) & Collision \\
		\midrule
            Sensorless & 135.2 & 59.1 & 0.51 & 0.0 \\
		Sensor-Based & 137.7 & 60.2 & 0.60 & 0.1 \\
		
		\bottomrule
        \end{tabularx}
	\label{tab:sensorless}
		\vspace{-0.2cm}
\end{table} 

\subsection{High-Speed Navigation}
\label{sec:app_highspeed}

\begin{table}[t]
	\centering
	\caption{ Result Comparison of High-Speed Navigation Scenario}  
	\begin{tabular}{c|cccc}
		\toprule
		Map & $d_{\rm{fly}}$(m) &  $t_{\rm{fly}}$(s) & $t_{\rm{cal}}$(ms) & Collision \\
		\midrule
            Proposed & 222.3 & 24.4 & 2.2 & 0.1 \\
		Robocentric & 230.2 & 33.8 & 2.6 & 0.5 \\
		\bottomrule
	\end{tabular}
	\label{tab:highspeed}
		\vspace{-0.2cm}
\end{table} 

In this scenario, a drone navigates through a simulation environment with velocity limit increased from \SI{3}{\mps} to \SI{9.3}{\mps}.
The sensor configuration in simulation is \SI{40}{\meter} range, \SI{60}{\degree} HFOV and \SI{45}{\degree} VFOV, to represent long range sensor such as solid-state LiDAR. The planning horizon is set to 25m. Test area is \SI{200}{\meter}$\times$\SI{35}{\meter}$\times$\SI{20}{\meter}. The proposed mapping system sets the parameters to $n_{lim}=50000$, $d_{in}=d_{inf}= \infty$, and the robocentric mapping system sets the boundary to \SI{16}{\meter}$\times$\SI{16}{\meter}$\times$\SI{16}{\meter} size for comparison. Both systems set resolution to \SI{0.1}{\meter}. The test involved 10 iterations. 

Fig. \ref{fig:application} (lower part) and Table \ref{tab:highspeed} shows the result of \textbf{1) the planner with proposed mapping system} and \textbf{2) the planner with array-based robocentric local map}. 
The test with proposed mapping system had maximum memory consumption for mapping of \SI{149}{\mb} and average planning computation time of \SI{2.2}{\ms}, while the test with robocentric local map had \SI{101}{\mb} and \SI{2.6}{\ms} for each. Navigation with proposed mapping system scored in average speed of \SI{9.1}{\mps} with \SI{90}{\percent} success rate. Navigation with robocentric map succeeded only \SI{50}{\percent} of the iterations, with average speed of \SI{6.8}{\mps}. 
The demonstration shows that our system can reinforce the planning algorithm's performance in high-speed, long-horizon navigation. The non-robocentric map of the proposed system effectively maintains awareness where it is needed, not only the robot's vicinity. 
The performance could be further improved, considering that we simply integrated the interface without changing feature of the planner or mapping system.

\section{Conclusion}
\label{sec:conclusion}

We proposed a voxel mapping system that can handle multiple mapping tasks on its own, including non-robocentric local mapping, long-range scanning, precise global map logging, and multi-agent mapping. 
Benchmark on multiple LiDAR datasets show that our system can achieve real-time performance, high resolution, and wide coverage simultaneously regardless of sensor and environment.
In addition, the proposed system includes efficient map-sharing feature for multi-agent mapping with significantly reduced overhead in network but no loss in map resolution. 
We integrated the proposed system into decentralized multi-agent planning system EGO-SWARM to demonstrate the applications that single-purposed maps can barely perform.

The proposed mapping system has far better versatility than other map algorithms and maintains real-time performance as well. However, it does not surpass other specialized maps in terms of pure performance. 
For the future work, we will try to further enhance the system by making it to automatically adjust the parameters in real-time based on system status, sensor data and task.
Also, we are working on communication-aware data managing framework, which can maintain decentralized network and share multi-agent information in a cluttered environment.

\bibliographystyle{reference/IEEEtran}
\bibliography{reference/main}

\end{document}